\documentclass[10pt,twocolumn,letterpaper]{article}

\usepackage[pagenumbers]{cvpr} 

\definecolor{cvprblue}{rgb}{0.21,0.49,0.74}
\usepackage[pagebackref,breaklinks,colorlinks,allcolors=cvprblue]{hyperref}

\def\paperID{3061} 
\def\confName{CVPR}
\def\confYear{2026}

\title{Neighbor GRPO: Contrastive ODE Policy Optimization Aligns Flow Models}

\author{
\bf Dailan He$^{1,2}$~~~~ Guanlin Feng$^{2}$~~~~ Xingtong Ge$^{2,3}$~~~~ Yazhe Niu$^{1}$~~~~ Yi Zhang$^{2}$ \\
\bf Bingqi Ma$^{2}$~~~~ Guanglu Song$^{2}$~~~~ Yu Liu$^{2}$~~~~ Hongsheng Li$^{1,4,5}$
\\ \\
$^1$CUHK MMLab ~~~ $^2$Vivix Group Limited ~~~ $^3$HKUST \\ $^4$Shenzhen Loop Area Institute ~~~ $^5$CPII under InnoHK \\
{\tt\small hedailan@link.cuhk.edu.hk ~~~~ hsli@ee.cuhk.edu.hk}
}

\usepackage{cancel}
\usepackage{algorithm2e}
\usepackage[dvipsnames, table]{xcolor}

\begin{document}
\maketitle
\begin{abstract}
  Group Relative Policy Optimization (GRPO) has shown promise in aligning image and video generative models with human preferences. However, applying it to modern flow matching models is challenging because of its deterministic sampling paradigm. Current methods address this issue by converting Ordinary Differential Equations (ODEs) to Stochastic Differential Equations (SDEs), which introduce stochasticity. However, this SDE-based GRPO suffers from issues of inefficient credit assignment and incompatibility with high-order solvers for fewer-step sampling. In this paper, we first reinterpret existing SDE-based GRPO methods from a distance optimization perspective, revealing their underlying mechanism as a form of contrastive learning. Based on this insight, we propose Neighbor GRPO, a novel alignment algorithm that completely bypasses the need for SDEs. Neighbor GRPO generates a diverse set of candidate trajectories by perturbing the initial noise conditions of the ODE and optimizes the model using a softmax distance-based surrogate leaping policy. We establish a theoretical connection between this distance-based objective and policy gradient optimization, rigorously integrating our approach into the GRPO framework. Our method fully preserves the advantages of deterministic ODE sampling, including efficiency and compatibility with high-order solvers. We further introduce symmetric anchor sampling for computational efficiency and group-wise quasi-norm reweighting to address reward flattening. Extensive experiments demonstrate that Neighbor GRPO significantly outperforms SDE-based counterparts in terms of training cost, convergence speed, and generation quality.
\end{abstract}

\begin{figure}
  \centering
  \includegraphics[width=0.9\linewidth]{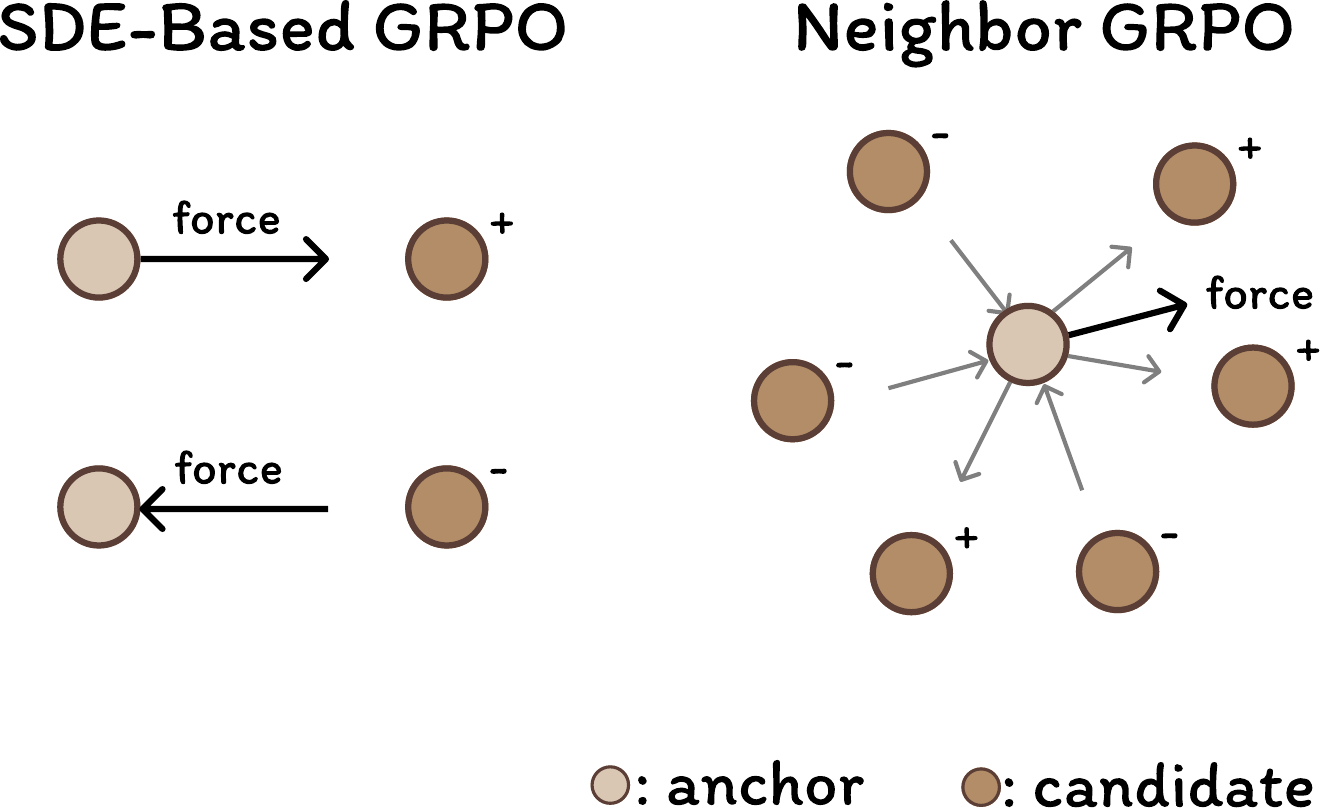}
  \caption{GRPO approaches for flow models optimize the sample $x_t$ at each timestep $t$. We revisit it from the perspective of contrastive learning, pushing \textit{anchor} samples to high-reward \textit{candidates} and vice versa. Different from SDE-based GRPO approaches~\cite{xue2025dancegrpo,liu2025flowgrpo} that conduct sample-wise exploration and optimization, our Neighbor GRPO optimizes the policy at the anchor in a joint force field defined by all candidates in a group. This approach allows full-ODE training, leading to better training efficiency and sample quality.}
  \label{fig:intro}
\end{figure}

\section{Introduction}
\label{sec:intro}

Building on the success of Reinforcement Learning from Human Feedback (RLHF)~\cite{christiano2017rlhf,ouyang2022rlhfinstruct,stiennon2020rlhf} and Reinforcement Learning with Verifiable Rewards (RLVR)~\cite{lambert2024tulu3,guo2025deepseekr1} in text generation, the application of Reinforcement Learning (RL) to align visual generative models with human preference is now a burgeoning field of research.
Among the various techniques, Group Relative Policy Optimization (GRPO)~\cite{shao2024deepseekmath} stands out as a powerful policy gradient method. By eliminating the need for an expensive critic network, GRPO achieves stable and efficient policy optimization, even with simple reward functions or rule-based verifiers. 
Specifically, the adaptation of GRPO to text-to-image diffusion models~\cite{ho2020ddpm,rombach2022ldm,ma2024exploring,zong2024easyref} and flow matching models~\cite{lipman2022flowmatching} has yielded promising results~\cite{xue2025dancegrpo,liu2025flowgrpo,li2025mixgrpo,li2025branchgrpo,wang2025prefgrpo}, with these methods showing superior performance in human preference alignment. Unlike previous alignment approaches~\cite{xu2023imagereward,wu2024drtune,deng2024flowdpo,wallace2024diffusiondpo,su2024ddpo}, those GRPO methods present stronger flexibility, requiring neither differentiable reward models nor paired human preference annotations.

State-of-the-art text-to-image generation models such as FLUX~\cite{labs2025flux1kontextflowmatching} have adopted a generation paradigm based on flow matching~\cite{lipman2022flowmatching}, which learns deterministic Ordinary Differential Equation (ODE) trajectories, defining from-noise-to-data probability flows. However, applying GRPO to flow matching models introduces a fundamental conflict: GRPO relies on stochasticity to explore the policy space, whereas the core advantage of flow matching models lies in their efficient, high-quality deterministic sampling.
Existing approaches, such as Flow-GRPO~\cite{liu2025flowgrpo} and DanceGRPO~\cite{xue2025dancegrpo}, primarily address this conflict by employing Stochastic Differential Equations (SDEs). This SDE-based GRPO approach converts the deterministic ODE into an equivalent SDE, modeling the reverse denoising process as a Markov decision process (MDP). 
This enables policy gradient-based RLHF optimization methods like GRPO. 

However, this conversion sacrifices the core advantages of ODEs. 
Firstly, SDE sampling is often restricted to first-order solvers, making it difficult to leverage more efficient high-order solvers like DPM-Solver++~\cite{lu2025dpm++}. 
Secondly, SDE training is relatively inefficient due to the mismatch in credit assignment between reward scores and single-step noise injection~\cite{li2025branchgrpo}. 
Subsequent works like MixGRPO~\cite{li2025mixgrpo} and BranchGRPO~\cite{li2025branchgrpo} have partially mitigated the computational efficiency issues through techniques like mixed sampling and branch reuse, but they remain constrained by the inherent limitations of the SDE framework and face the challenge of credit assignment, \ie the terminal reward signal must be distributed across all timesteps. Hence, a novel paradigm is urgently needed.

We re-examine the optimization mechanism of GRPO-based flow matching from the perspective of inter-sample distance optimization. 
For existing SDE-based GRPO schemes, at each sampling step, we can reinterpret the optimization dynamics: if we consider the deterministically computed sample from the ODE as an \textit{anchor} and the stochastically sampled counterparts from the SDE as \textit{candidates}, optimizing the SDE policy gradient is equivalent to pulling the anchor closer to high-reward candidates while pushing it away from low-reward ones. This dynamic is illustrated in Figure~\ref{fig:intro}.
Therefore, the SDE-based GRPO optimization process can be understood as a form of \textbf{contrastive learning}~\cite{chopra2005contrastiveloss,schroff2015facenet,frosst2019softnn,he2020moco} based on the distance between the anchor and candidate samples. 
Since the candidate samples obtained from SDE sampling are random perturbations of the ODE anchor, this contrastive learning process guides the model to explore positive and negative samples in the neighborhood and determine the optimization direction.

From this perspective, we propose a more direct approach: we generate a set of candidate trajectories by perturbing the initial conditions of the ODE, effectively creating a neighborhood around the original ODE path. We then select one trajectory as the anchor and the others as candidates, applying a distance-based loss to pull the anchor towards high-reward candidates and push it away from low-reward ones. Inspired by contrastive learning, we adopt an advantage-guided soft nearest-neighbor optimization to achieve this distance optimization. 
The key finding is that the softmax distribution corresponding to this objective defines a \textbf{surrogate leaping policy} in the ODE neighborhood. 
This policy introduces the necessary stochasticity for exploring the deterministic ODE process by performing single-step random transitions on neighborhood samples according to the softmax distribution of anchor-candidate distances. 
Based on this surrogate policy, we can rigorously integrate distance optimization directly into the GRPO framework, allowing the entire optimization process to break free from its dependence on SDE sampling and achieve effective reinforcement learning while fully preserving all the advantages of ODEs. We further investigate key points for practical implementation: \textbf{(1)} a symmetric anchor sampling strategy to accelerate policy updates; \textbf{(2)} the direct use of high-order solvers like DPM++~\cite{lu2025dpm++}; and \textbf{(3)} a quasi-norm reweighting mechanism to address reward hacking by sharpening the often-flat GRPO advantage vector. Both qualitative and quantitative results demonstrate that our proposed approach outperforms existing methods in text-to-image generation with only 4 hours of training.

In this paper, we propose Neighbor GRPO, a novel RLHF method for flow matching models. Our main contributions include:
\begin{enumerate}
  \item We reinterpret GRPO in flow matching from a distance optimization perspective, revealing its connection to contrastive learning. 
  \item We propose Neighbor GRPO based on a discrete surrogate policy, which constructs a contrastive learning objective in the ODE neighborhood. This achieves effective policy optimization while fully preserving the advantages of ODE solvers.
  \item We demonstrate that Neighbor GRPO can outperform existing SDE-based methods in terms of training cost and generation quality through extensive experiments.
\end{enumerate}

\section{Related Works}


The application of RLHF to visual generative models has evolved rapidly, with GRPO~\cite{shao2024deepseekmath} emerging as a particularly effective approach due to its stability and elimination of expensive critic networks. We categorize recent advances into two waves: pioneering GRPO adaptations designed for flow models and efficiency improvements for practice.

Flow-GRPO~\cite{liu2025flowgrpo} and DanceGRPO~\cite{xue2025dancegrpo} successfully integrate online RL with flow matching models by converting the deterministic ODE into an equivalent SDE, enabling stochastic exploration while maintaining marginal probability distributions. 
DanceGRPO extends this approach to a broader range of visual generation tasks (text-to-image, text-to-video, image-to-video) and emphasizes key implementation details such as shared noise initialization to prevent reward hacking.
MixGRPO~\cite{li2025mixgrpo} introduces a hybrid ODE-SDE sampling strategy with a sliding window mechanism. By restricting SDE sampling and GRPO optimization to a window that slides from high-noise to low-noise regions, it significantly reduces training costs while achieving better performance.
BranchGRPO~\cite{li2025branchgrpo} tackles efficiency and credit assignment from a different angle by replacing independent sequential rollouts with a branching tree.

Our work builds upon these advances but takes a fundamentally different approach. While existing methods rely on SDE conversion, we propose a novel perspective that reinterprets SDE-based GRPO as distance-based contrastive learning and introduces a surrogate leaping policy that enables effective RLHF while preserving all the advantages of deterministic ODE sampling.

\section{Methodology}

In this section, we present Neighbor GRPO, a novel RLHF method for flow matching models that bypasses the need for SDE conversion. We begin by establishing necessary background on flow matching and SDE-based GRPO methods (Section~\ref{sec:prelim}). We then reinterpret existing SDE-based GRPO approaches from the perspective of distance-based contrastive learning (Section~\ref{sec:sde-contrastive}). Building on this insight, we introduce Neighbor GRPO (Section~\ref{sec:neighbor-grpo}), which constructs a neighborhood of candidate trajectories by perturbing initial noise conditions and defines a softmax distance-based surrogate leaping policy to enable policy gradient optimization while preserving deterministic ODE sampling. Finally, we present three key implementation strategies of our proposed Neighbor GRPO (Section~\ref{sec:practical}).

\subsection{Preliminary}
\label{sec:prelim}

\paragraph{Flow Matching Models.}

Flow matching models~\cite{lipman2022flowmatching} generate samples by learning deterministic Ordinary Differential Equation (ODE) trajectories from noise to data. Its forward and backward processes are defined as
\begin{equation}
x_t = (1-t)x + t\epsilon, \quad t \in [0,1],
\end{equation}
\begin{equation}
x_{t-\Delta t}^{(\mathrm{ODE})} = x_t - \Delta t v_\theta(x_t,t), \quad \Delta t > 0,
\label{eq:fm-step}
\end{equation}
where $x$ is a data sample and $\epsilon \sim \mathcal{N}(0,I)$ is standard Gaussian noise. The model learns a velocity field $v_\theta(x_t, t)$ such that the reverse denoising process can be solved similarly to v-prediction diffusion models.
This deterministic property gives flow matching a significant advantage in sampling efficiency but also creates a fundamental conflict with the stochastic exploration required by reinforcement learning.

\paragraph{SDE-Based GRPO Methods.}

Let $s_t = (c, t, x_{t+\Delta t})$ be the state at timestep $t$, where $c$ represents the generation conditions. GRPO~\cite{shao2024deepseekmath}, as one of the policy gradient methods~\cite{sutton1999reinforce,schulman2015trpo,schulman2017ppo}, directly optimizes the parameters $\theta$ of a policy $\pi_\theta(x_t|s_t)$, which is the conditional probability distribution defined by the generative model. A reward function $R(x_0; c)$, given only at the terminal timestep, defines the supervision reward $r$. GRPO defines a group-wise normalized advantage function, computed from $G$ rollout samples:
\begin{equation}
A_i = \frac{r_i - \text{mean}(\{r_1, \ldots, r_G\})}{\text{std}(\{r_1, \ldots, r_G\})}.
\label{eq:grpo-norm}
\end{equation}
Similar to PPO~\cite{schulman2017ppo} using clipped policy ratio between old and on-the-fly policies, the GRPO objective is written as
\begin{equation}
\mathcal{J}(\theta) = \mathbb{E}_{s, t, i} \left[  \min\left( A_i \rho_{t}^{(i)} , A_i \lceil\rho_{t}^{(i)}\rfloor \right) \right] - \beta D_{\text{kl}}[\pi_{\theta_{\text{old}}} \| \pi_{\theta} ],
\label{eq:Jgrpo}
\end{equation}
where $\rho_{t}^{(i)} = \frac{\pi_\theta(x_{t}^{(i)}|s_{t}^{(i)})}{\pi_{\theta_{\text{old}}}(x_{t}^{(i)}|s_{t}^{(i)})}$ is the policy ratio and $\lceil\cdot\rfloor$ denotes ratio clipping. It is popular in recent studies to remove the KL-divergence term~\cite{xue2025dancegrpo,li2025mixgrpo}, since the ratio clip with a tiny threshold ($10^{-4}$ in DanceGRPO and MixGRPO) strongly ensures the parameters in the trust region.

To introduce stochasticity into the deterministic ODE framework to support policy gradient optimization, SDE-based GRPO methods~\cite{liu2025flowgrpo,xue2025dancegrpo} adopt SDE sampling with the same marginal probability density as the original ODE:
\begin{equation}
x^\mathrm{(SDE)}_{t-\Delta t} = x^\mathrm{(ODE)}_{t-\Delta t} -  \frac{\sigma_t^2}{2t} \hat{\epsilon}_\theta(x_t,t)    +  \sigma_t \epsilon,
\end{equation}
where $\hat{\epsilon}_\theta = x_t + (1-t)v_\theta(x_t,t)$ is the estimated initial noise according to flow matching (Eq.~\ref{eq:fm-step}). It illustrates the drift term of the SDE. $\sigma_t$ is a factor controlling the strength of stochasticity, and $\epsilon \sim \mathcal{N}(0,I)$ is independent Gaussian noise. We adopt a formulation that is slightly different from but equivalent to that in previous literature, and we provide a derivation in the supplementary material. 
Under this framework, the policy $\pi_\theta(x_{t}|s_{t})$ becomes an isotropic Gaussian distribution of $x_t$ conditioned on $x_{t+\Delta t}$ whose likelihood can be explicitly computed in an exp-MSE form, thus supporting optimization of the GRPO objective.

\subsection{SDE-Based GRPO Is Secretly Contrastive Learning}
\label{sec:sde-contrastive}

To better understand the optimization mechanism of SDE-based GRPO, we reexamine its optimization dynamics from a distance optimization perspective. This analysis will lay the theoretical foundation for our proposed method.

The policy ratio objective is referred to as the Conservative Policy Iteration (CPI) objective~\cite{kakade2002cpi,schulman2015trpo,schulman2017ppo,shao2024deepseekmath}. It has been proven to be an estimation of the vanilla policy gradient or REINFORCE objective~\cite{sutton1999reinforce}, which is a log-likelihood expectation $\mathbb{E}[A_i \nabla_\theta \log \pi_\theta(x_t^{(i)} \mid s_t^{(i)})]$. The error is bounded by the policy difference $\theta - \theta_\text{old}$~\cite{kakade2002cpi,schulman2015trpo}. The core of this log-policy term in SDE-based GRPO is the negative squared distance. Since the policy ratio is clipped into the trust region~\cite{schulman2015trpo, schulman2017ppo}, we have $\theta \approx \theta_{\text{old}}$.
Therefore, maximizing the SDE-based GRPO objective is equivalent to minimizing an advantage-weighted MSE loss:
\begin{equation}
  -\log\pi_\theta = \frac{1}{2\sigma_{t+\Delta t}^2} \left\| \tilde{x}_{t}^{(\mathrm{ODE})} - x_{t}^{(\mathrm{ODE})} + o_t(x_{t+\Delta t}) \right\|^2_2,
\end{equation}
where the constant term is omitted, and $\tilde{x}_{t}^{(\mathrm{ODE})} = x_{t}^{(\mathrm{ODE})} + \sigma_{t+\Delta t}\epsilon$ is the perturbed ODE sample. The $o_t(x_{t+\Delta t})$ term denotes a drift residual.
As the policy $\theta$ is constrained to the trust region of $\theta_{\text{old}}$~\cite{schulman2015trpo,schulman2017ppo}, this residual term $o(\cdot)$ approaches zero and contributes minimally to the optimization. We present the full derivation in the supplementary material.

This form reveals the core optimization mechanism of SDE-based GRPO: advantage-weighted policy gradient updates are equivalent to making the ODE sample approximate nearby stochastically perturbed targets. See Figure~\ref{fig:intro}. This is a distance optimization process. With advantage weighting, high-reward samples (positive $A_i$) are pulled closer, and low-reward samples (negative $A_i$) are pushed away, essentially forming a contrastive learning~\cite{chopra2005contrastiveloss,schroff2015facenet,frosst2019softnn,he2020moco} mechanism centered on the one-step ODE trajectory. 

\begin{figure}
  \centering
  \includegraphics[width=0.9981\linewidth]{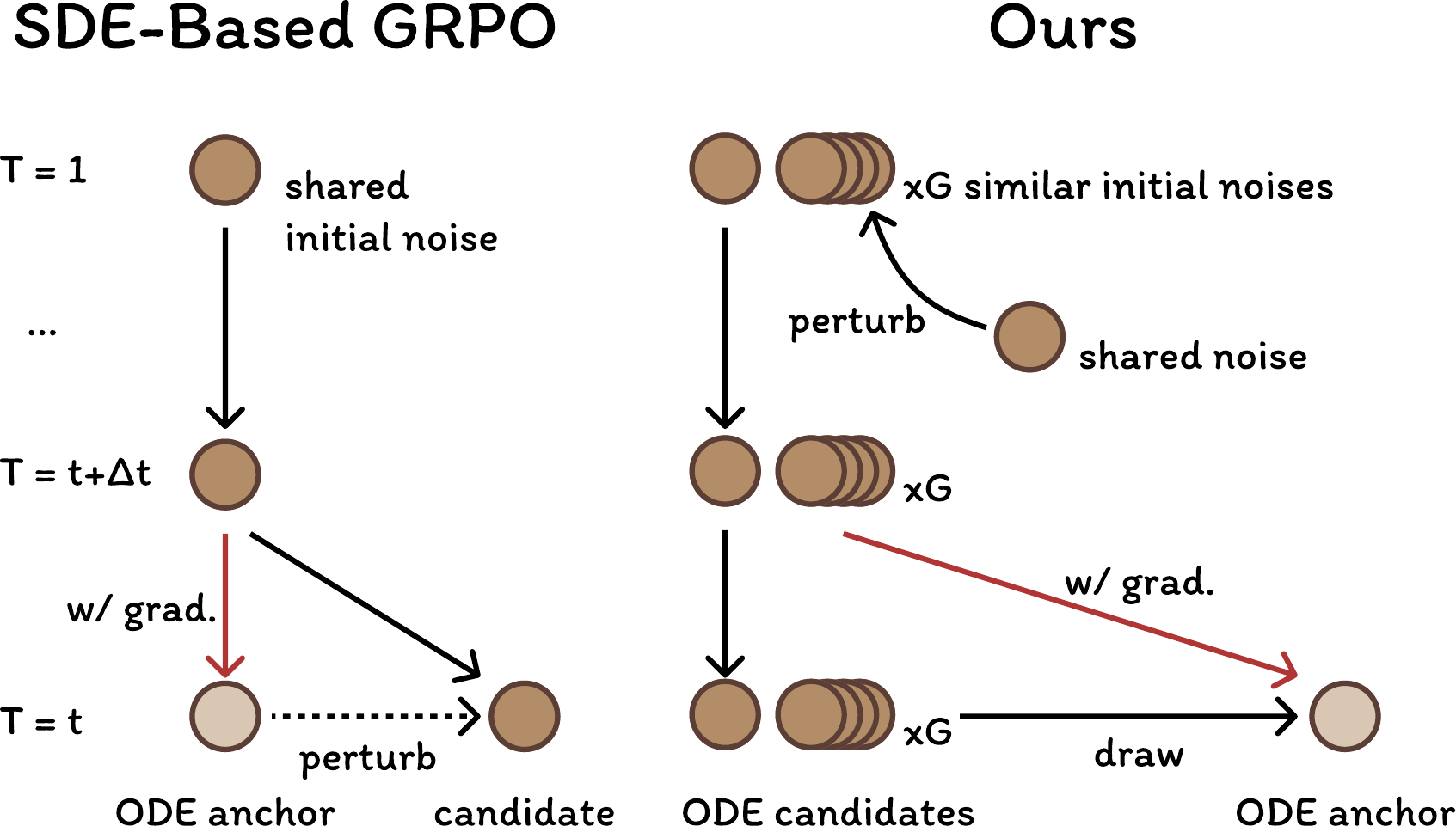}
  \caption{Different from SDE-based GRPO approaches, which explore the sample space with noise perturbation defined by SDE, we directly construct a group of similar initial noises, and conduct deterministic ODE sampling.}
  \label{fig:sample}
\end{figure}

\subsection{Neighbor GRPO}
\label{sec:neighbor-grpo}

SDE-based GRPO can be viewed as contrastive learning: ODE anchors are pulled toward high-reward SDE samples and pushed away from low-reward ones. Since candidates act only as reference points and do not backpropagate, we ask: can we eliminate SDEs and optimize a distance objective directly in the ODE neighborhood?

We propose a method following this intuition. Instead of using SDEs to generate stochastic variations for exploration, we create a set of candidate trajectories by
perturbing the initial noise of the deterministic ODE solver (Figure~\ref{fig:sample}). Given a base initial noise $\epsilon^*$, we construct $G$ initial conditions:
\begin{equation}
\epsilon^{(i)} = \sqrt{1-\sigma^2} \epsilon^* + \sigma \delta^{(i)}, \quad i=1,\dots,G,
\label{eq:init-noise-perturb}
\end{equation}
where $\delta^{(i)} \sim \mathcal{N}(0, I)$ and $\sigma \in (0,1)$ controls the perturbation strength. Evolving these conditions produces a bundle of trajectories forming a local solution neighborhood. 

To fit GRPO, a stochastic policy $\pi_\theta(x_t|s_t)$ among those trajectories is required, but ODE sampling is deterministic. We therefore introduce a training-only \textit{surrogate policy} that enables policy ratios and gradients while keeping inference fully deterministic.

\vspace{-0.5cm}\paragraph{Softmax-Distance Is a Natural Choice.}
This surrogate should (1) favor candidates closer to the anchor, (2) be tractable for policy gradients, and (3) align with contrastive learning. The softmax distribution over negative squared distances meets these criteria, paralleling soft nearest neighbor~\cite{frosst2019softnn} and InfoNCE~\cite{oord2018infonce}. We define the \textit{surrogate leaping policy} over $G$ candidates in a GRPO batch at timestep $t$:
\begin{align}
  \pi_\theta\left(x_t^{(i)} \mid \{s_t\} \right) 
  &= \frac{\exp\big(-\|x_t^{(i)} - x_t^{(\theta)}\|_2^2\big)}{\sum_{k=1}^G \exp\big(-\|x_t^{(k)} - x_t^{(\theta)}\|_2^2\big)},
  \label{eq:leap-policy}
\end{align}
where $\{s_t\} = \{s_t^{(1)}, s_t^{(2)}, \dots, s_t^{(G)}\}$ is the group state. The anchor $x_t^{(\theta)}$ is randomly drawn from the candidates $\{x_t\}$, and it contributes gradients to the parameters $\theta$. This distribution typically assigns the highest probability to the nearest candidate, which is often the anchor itself.

\begin{figure}[t]
  \centering
      \centering
      \includegraphics[width=0.95\linewidth]{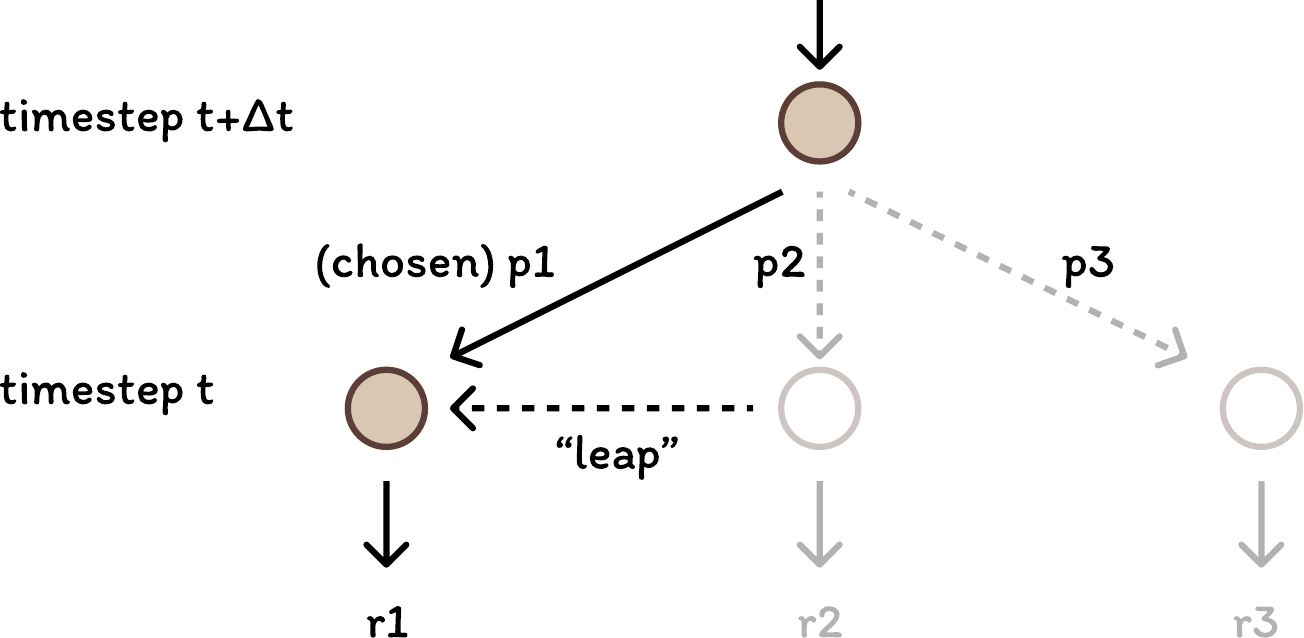}
  \caption{Surrogate leaping policy. The ongoing sampling trajectory virtually leaps to another, following the policy distribution: $p_i = \pi_\theta\left(x_t^{(i)} \mid \{s_t\} \right)$}
  \label{fig:leap}
\end{figure}

\vspace{-0.5cm}\paragraph{Virtual Leaping in the ODE Neighborhood.}
We interpret $\pi_\theta$ as a virtual stochastic process over the ODE neighborhood (Figure~\ref{fig:leap}): at each $t$, the trajectory may ``leap'' to a neighbor with probability $\pi_\theta$, most often staying on the anchor. The surrogate policy approximately preserves the ODE marginal (see supplementary material for derivation).

This surrogate is an \textit{implicit training tool}, not an inference mechanism. During training it provides policy ratios and gradients that bias the anchor toward high-reward regions; at inference we run standard deterministic ODE sampling without leaping or softmax computation.

\vspace{-0.5cm}\paragraph{Optimization Dynamics.}
Maximizing its policy gradient via the GRPO objective (Eq.~\ref{eq:Jgrpo}) yields a simple geometry: for $A_i>0$, the gradient decreases the distance $\|x_t^{(i)}-x_t^{(\theta)}\|_2^2$, pulling the anchor toward $x_t^{(i)}$; for $A_i<0$, it pushes the anchor away. Thus, distance-based contrastive learning is realized within a policy-gradient framework while preserving deterministic ODE inference.

\subsection{Practical Implementations}
\label{sec:practical}

In this section, we introduce three key implementation strategies that significantly enhance the method's practical effectiveness: (1) \textit{symmetric anchor sampling} that exploits the neighborhood structure to reduce the number of backward passes per iteration; (2) a \textit{group-wise quasi-norm advantage reweighting} mechanism that addresses the challenge of reward flattening in later training stages, preventing the model from being trapped in local optima and ensuring continued improvement; (3) usage of \textit{high-order solvers} such as DPM++, enabling faster sampling and better generation quality.

\subsubsection{Symmetric Anchor Sampling}

Due to the Johnson–Lindenstrauss Lemma~\cite{johnson1984JLlemma}, the initial noises in a sample group (as formulated in Eq.~\ref{eq:init-noise-perturb}) are almost equidistant. This provides a symmetry of the samples: each one is a ``center" for the others.
Due to this symmetry, any candidate in the sample group is a potential choice of anchor. 
This symmetric anchor sampling strategy enables a reduction in the number of required anchors during one GRPO iteration.
Concretely, the GRPO objective for a drawn anchor with index $k$ at time $t$ is
\begin{equation}
  \sum_{i=1}^G \min\!\left( A_i \,\rho_{t}^{(i\mid k)},\, A_i \left\lceil \rho_{t}^{(i\mid k)} \right\rfloor \right),
  \label{eq:anchor-merging-motivation}
\end{equation}
where $\rho_{t}^{(i\mid k)}$ is the anchor-conditioned policy ratio:
\begin{equation}
\rho_{t}^{(i\mid k)} = \frac {\pi_\theta(x_t^{(i)} \mid \{s_t\})} {\pi_{\theta_\text{old}}(x_t^{(i)} \mid \{s_t\} )} \Bigg|_{x_t^{(\theta)} = x_t^{(k)}}.
\end{equation}
In the surrogate leaping policy (Eq.~\ref{eq:leap-policy}), since all candidate samples $x_t^{(i)}$ are generated by the old policy and only the anchor path $x_t^{(\theta)}$ depends on $\theta$, computing this objective and its gradient requires a single forward–backward pass per GRPO batch. Note that existing SDE-based GRPO approaches~\cite{xue2025dancegrpo, li2025mixgrpo} are typically sample-wise and require $G$ such calculations instead. For commonly used FLUX training with $G=12$, our method saves up to 12 times of the forward-backward calculations during policy update. This is a strong advantage of our proposed group-wise method.

In practice, because of the symmetry, we sample $B$ anchors per GRPO iteration to estimate a more accurate gradient. Since $B < G$, this few-anchor speed-up still benefits training by balancing performance and computational cost.

\subsubsection{Group-Wise Quasi-Norm Advantage Reweighting}
\begin{figure}[t]
    \centering
    \begin{subfigure}{0.3\linewidth}
        \centering
        \includegraphics[width=0.99\linewidth]{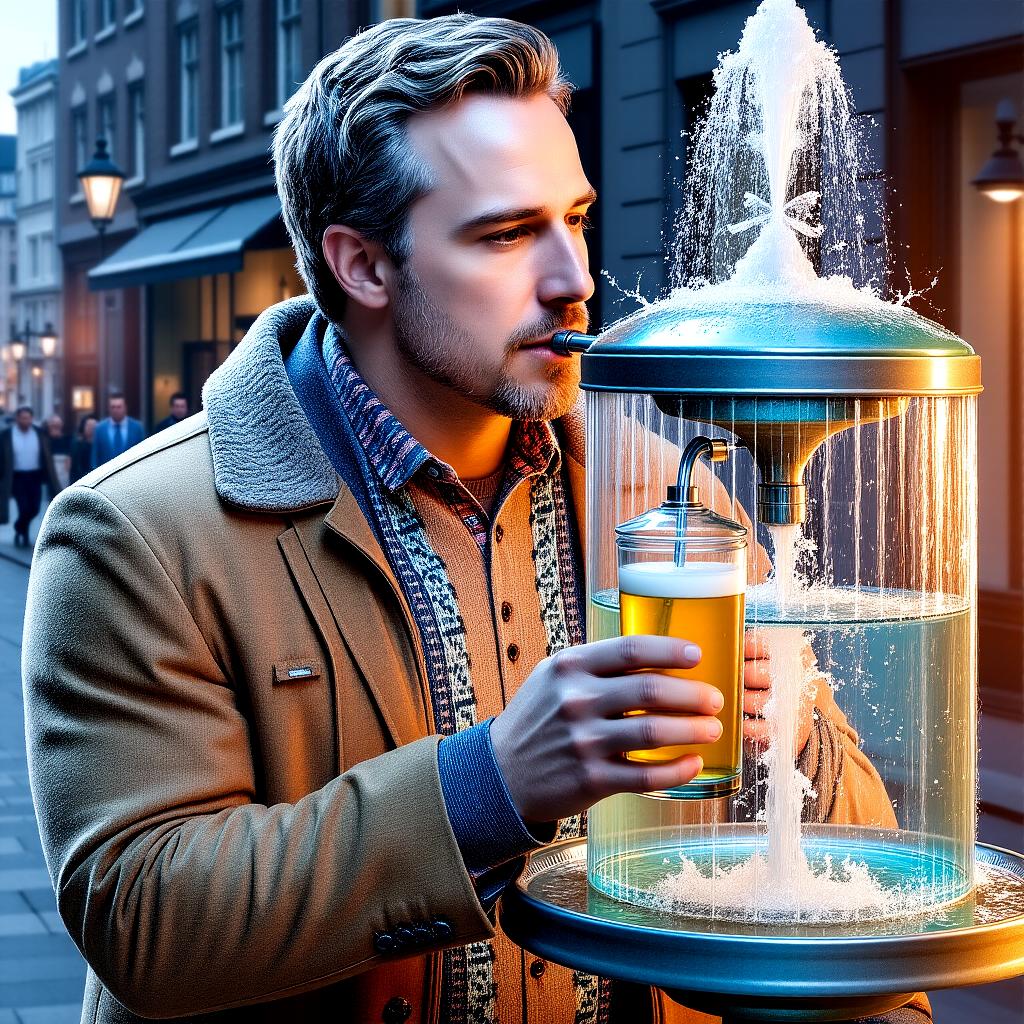}
        \caption{$p=2$ (GRPO)}
        \label{fig:pnorm_intro:std}
    \end{subfigure}
    \begin{subfigure}{0.3\linewidth}
        \centering
        \includegraphics[width=0.99\linewidth]{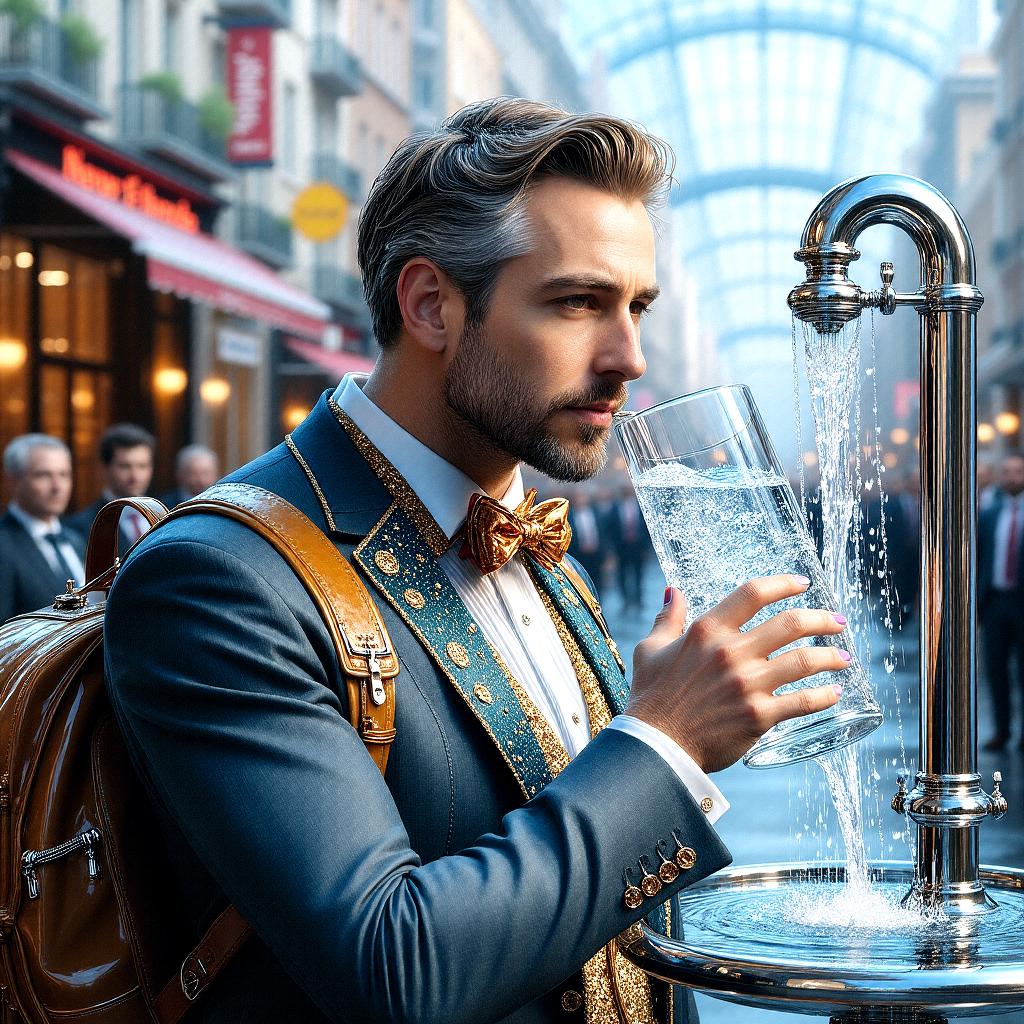}
        \caption{$p=1$}
    \end{subfigure}
    \begin{subfigure}{0.3\linewidth}
        \centering
        \includegraphics[width=0.99\linewidth]{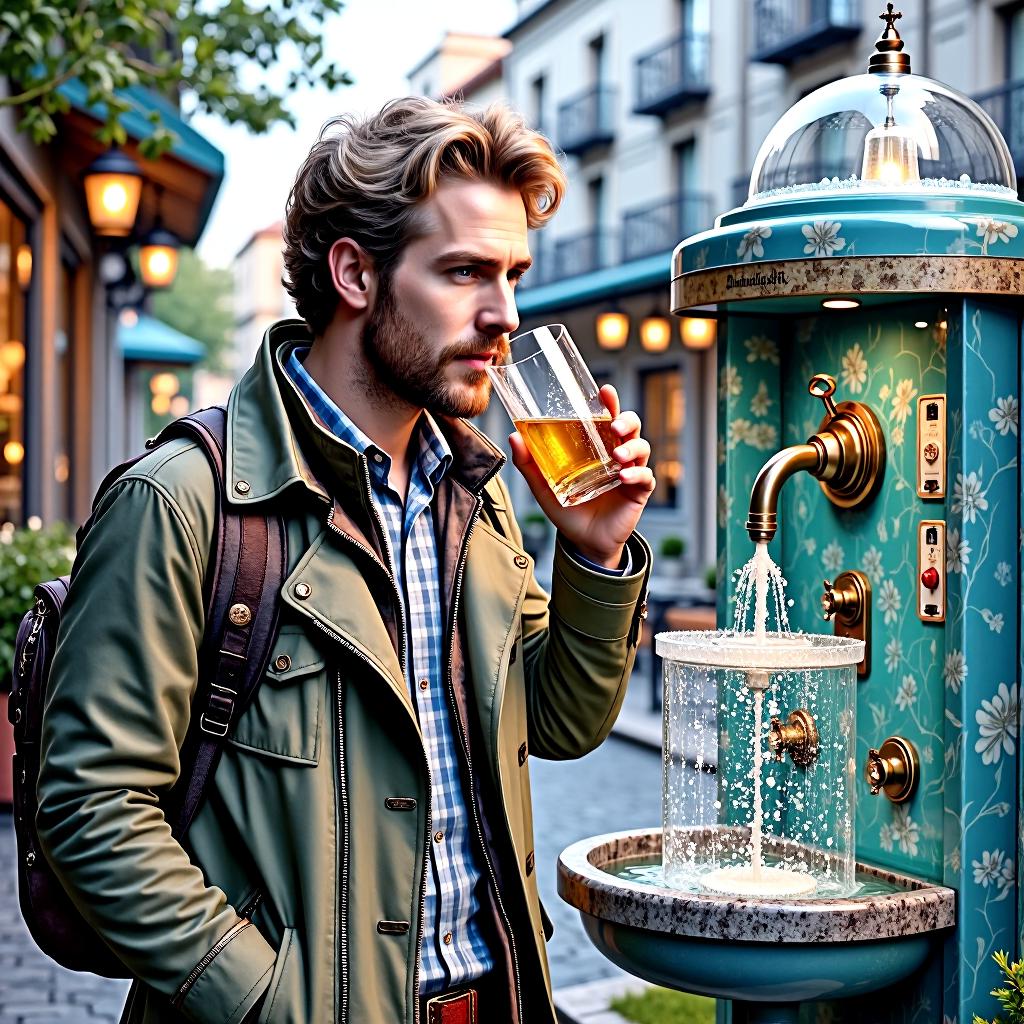}
        \caption{$p=0.8$}
    \end{subfigure}
    \caption{Effect of quasi-norm reweighting.}
    \label{fig:pnorm_intro}
\end{figure}

A key challenge in our neighborhood-based optimization arises when the advantage signal flattens after sufficient optimization. The softmax mechanism in our surrogate policy relies on advantage-weighted distances to distinguish high-reward from low-reward candidates. However, when all samples in a group yield similar rewards, the standard GRPO normalization (Eq.~\ref{eq:grpo-norm}) produces a nearly uniform advantage distribution. Although these advantages are numerically small, they still guide the optimization, leading to weak, over-averaged updates that can cause reward hacking artifacts such as generic average faces in portrait generation (Figure~\ref{fig:pnorm_intro:std}). This issue of flattened rewards is a known challenge in GRPO and related methods~\cite{liu2025drgrpo,yu2025dapo}.

Standard GRPO normalization projects the advantage vector $\mathbf{A} = (A_1, \ldots, A_G)$ onto an $L_2$-sphere of fixed radius, treating all groups equally regardless of the internal structure of their reward signals. We propose to replace this with a group-wise quasi-normalization based on the $L_p$-norm with $p < 2$:
\begin{equation}
A'_i = \frac{A_i}{\left(\sum_{k=1}^G |A_k|^p\right)^{1/p}}, \quad p \in (0,2].
\end{equation}
The insight is that quasi-norms with $p<2$ exhibit sparsity-inducing behavior. For a flat advantage vector where all $|A_k|$ are approximately equal, the $L_p$-norm grows approximately as $G^{1/p}$, which is large when $p$ is small. Dividing by this large norm shrinks the rescaled advantages $A'_i$, effectively down-weighting the contribution of this uninformative group to the overall gradient. When $p=2$, the $L_2$-norm becomes independent of the sparsity pattern (up to variance), recovering the standard GRPO normalization.

This reweighting mechanism is compatible with existing GRPO clipping schemes and introduces negligible computational overhead. Crucially, it preserves the sign and relative ordering of advantages within each group, maintaining the contrastive learning dynamic while adaptively emphasizing groups that provide clear optimization signals.

\subsubsection{Usage of High-Order Solvers}
Decoupling full-ODE sampling from the distance-based policy enables us to adopt high-order solvers such as DPM++~\cite{lu2025dpm++} to improve both training efficiency and final performance. Concretely, we roll out all trajectories with DPM++ for data collection, while during policy updates we use a one-step DDIM~\cite{song2020ddim} transition from $x_{t+\Delta t}$ to $x_t$ to compute the surrogate policy and the GRPO objective.

\newcommand{\official}{$^\dag$}
\newcommand{\DA}{{$\downarrow$}}
\newcommand{\UA}{{$\uparrow$}}


\definecolor{DEEPGREEN}{HTML}{7FC8C9}
\definecolor{LIGHTGREEN}{HTML}{AFE0DE}
\definecolor{SHALLOWGREEN}{HTML}{D7F0EF}

\newcommand{\mybox}[2]{\par\noindent\colorbox{#1}
{{#2}}}

\newcommand{\Topfst}[1]{{\cellcolor{DEEPGREEN}{#1}}}
\newcommand{\Topscnd}[1]{{\cellcolor{LIGHTGREEN}{#1}}}
\newcommand{\Topthrd}[1]{{\cellcolor{SHALLOWGREEN}{#1}}}

\begin{table*}[t]
    \centering
\scalebox{0.9}{
    \begin{tabular}{llrrrrrrrrr}
    \toprule
    \textbf{Method} & \textbf{Solver}$_{\text{old}}$ & \textbf{NFE}$_{\text{old}}$\DA & 
    \textbf{NFE}$_{\theta}$\DA   & 
    \textbf{s / Iter.}\DA &
    \textbf{HPSv2.1}\UA  & \textbf{Pick}\UA & \textbf{Img.Rwd.}\UA & \textbf{CLIP}\UA & \textbf{Unified.}\UA & \textbf{Aes.}\UA  \\
    \midrule
    \official FLUX.1-dev      & N/A &  N/A & N/A & N/A & 0.310  & 0.227 & 1.131 &  \Topscnd{0.389} & 3.211 & 6.108\\
    \midrule
    \official DanceGRPO & DDIM & 25 & 14 & 237.86& \Topfst{\underline{0.371}}   &          0.231 &          1.306 &   \underline{0.364} & 3.156 & 6.552\\
    \official MixGRPO   & DDIM & 25 &  14  & 237.71 & \underline{0.366}   & \Topscnd{\underline{0.235}} & \Topthrd{\underline{1.604}} &            0.382 & 3.257 & 6.623 \\
    \quad-Flash* & DPM++ & 8 & 4 & 129.36 & \underline{0.343} & \underline{0.228} & \underline{1.409} & 0.374 & 3.148 & \Topfst{6.770}\\
    \midrule
    NeighborGRPO & DDIM & 25 & 4 & 141.93 & \underline{0.359}   & {\underline{0.234}} & \underline{1.571} &            0.385 & \Topthrd{3.262} & \Topscnd{6.669}\\
    NeighborGRPO &DPM++ & 16 & 3 & 105.85  & \Topscnd{\underline{0.369}}   & \Topfst{\underline{0.236}} & \Topfst{\underline{1.643}} & \Topthrd{0.386} & \Topscnd{3.324} & \Topthrd{6.633} \\
    NeighborGRPO & DPM++ & 8 & 1.33 & 45.08 & \Topthrd{\underline{0.366}}   & \Topthrd{\underline{0.234}} & \Topscnd{\underline{1.640}} & \Topfst{0.391} & \Topfst{3.334} & 6.621\\
    
    \bottomrule
    \end{tabular}
}
    \caption{Comparison of human preference scores. \underline{Underline}: in-domain preference. $\dag$: official checkpoints.}
    \label{tab:major_cmp}
\end{table*}

\section{Experimental Results}

We closely follow existing protocols~\cite{xue2025dancegrpo,li2025mixgrpo} to conduct training and evaluations for a fair comparison. We choose FLUX.1-dev~\cite{labs2025flux1kontextflowmatching} as the base model to evaluate all the algorithms. We adopt the HPDv2 training and test sets, which is an official dataset used by HPSv2~\cite{wu2023hpsv2}. It includes 103,700 prompts, from which 4800 or 9600 are randomly sampled for training over 300 iterations with a batch size of 16 or 32. The full 3200-sample test set is adopted to evaluate each model. We train each model with the AdamW optimizer for 300 iterations, with a learning rate of $10^{-5}$ and a batch size of 1 per GPU. For ablation studies, we use 16 GPUs to reduce the cost. For the model performance evaluation and comparison in Table~\ref{tab:major_cmp}, we use 32 GPUs, following existing protocols~\cite{xue2025dancegrpo, li2025mixgrpo}. In single reward training and ablation studies, we adopt the HPSv2.1 score~\cite{wu2023hpsv2} as the optimization target. Following MixGRPO, we evaluate multi-reward training with HPSv2.1, Pick Score~\cite{kirstain2023pick} and ImageReward~\cite{xu2023imagereward}, because this joint training is demonstrated to be effective against reward hacking~\cite{li2025mixgrpo,xue2025dancegrpo}. We evaluate the models using a multi-dimensional benchmark to cover both in-domain and out-of-domain assessments, consisting of CLIP Score~\cite{ilharco_gabriel_2021_5143773}, LAION Aesthetic Score~\cite{schuhmann2022laion}, and UnifiedReward~\cite{unifiedreward}. We also report the number of function
evaluations (NFEs)~\cite{lu2022dpm} to assess the training cost.

\subsection{Comparison}
We select DanceGRPO~\cite{xue2025dancegrpo} and MixGRPO~\cite{li2025mixgrpo} as baselines, and compare our Neighbor GRPO with them. They stand for pure SDE-based approaches and SDE-ODE mixed methods, respectively. We also reproduce recently proposed BranchGRPO~\cite{li2025branchgrpo} for reference.

\begin{table}[t]
    \centering
\scalebox{0.9}{
    \begin{tabular}{lrrrrrr}
    \toprule
    \textbf{Method}  &  \underline{\textbf{HPSv2.1}}  & \textbf{CLIP} & \textbf{Unified.} & \textbf{Aes.}  \\
    \midrule
    DanceGRPO & 0.356 & 0.366 & 3.114 & 6.400\\
    MixGRPO   & 0.371 & 0.367 & 3.125 & 6.365\\
    BranchGRPO & 0.379 & 0.362 & 3.138 & 6.571\\
    \midrule
    Ours $p=2$  & \textbf{0.379} & 0.368 & 3.132 & 6.503 \\ 
    Ours $p=0.8$ & 0.372 & \textbf{0.371} & \textbf{3.166} &  \textbf{6.626}\\
    \bottomrule
    \end{tabular}
}
    \caption{Various approaches optimized toward HPSv2.1.}
    \label{tab:hps_cmp}
\end{table}
\begin{figure}[t]
    \centering
    \includegraphics[width=0.9\linewidth]{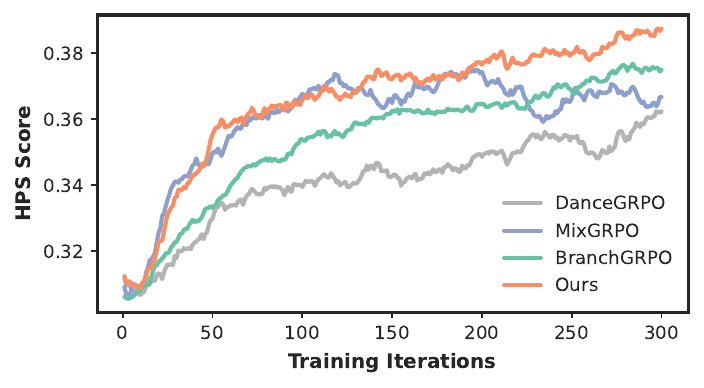}
    \caption{Training curves towards HPSv2.1.}
    \label{fig:hps_train_curves_cmp.tex}
\end{figure}

\vspace{-0.3cm}\paragraph{Single Target.}
See Table~\ref{tab:hps_cmp}. When training with HPSv2.1, our approach strikes a balance between training efficiency and final performance, with remarkable performance on both in-domain (HPSv2.1) and out-of-domain evaluations.

As presented in Figure~\ref{fig:hps_train_curves_cmp.tex}, our approach converges faster than the DanceGRPO baseline, reaching an HPSv2.1 score of over $0.35$ with only 50 iterations. It also achieves better training stability than MixGRPO in long-term training, implying a possibility for further scaling up.

\paragraph{Multiple Targets.}
To achieve the best visual quality, we follow existing protocols to train the model with multiple rewards. See Table~\ref{tab:major_cmp}. We apply the same multi-reward objective as MixGRPO, which adopts HPSv2.1, Pick Score, and Image Reward with equal weights. In 25-step DDIM rollout experiments, our model achieves the highest scores on all out-of-domain evaluations, while remaining on par with the baselines on in-domain evaluations. We believe this high out-of-domain performance is a positive sign, implying robustness against reward hacking and demonstrating the superiority of our approach. In few-step DPM++ solver experiments comparing to MixGRPO-Flash, our approach achieves remarkable performance in most metrics with both 16-step and 8-step rollouts. This proves our model is effective in few-step DPM++ training, demonstrating the superiority of our full-ODE scheme.

\vspace{-0.3cm}\paragraph{Training Efficiency.}
We analyze the training efficiency in Table~\ref{tab:major_cmp}, and list more details in Table~\ref{tab:train_cost}. Thanks to the symmetric anchor sampling strategy,
our approach also requires fewer forward/backward calculations during policy update. 
Given that $B$ is the number of anchors sampled in one GRPO iteration, and $G$ denotes the GRPO group size. Let $K$ denote how many timesteps are updated in a GRPO iteration, usually related to the parameter \texttt{timestep\_fraction}. The training cost during policy update is $O(BK)$.
Since the baseline methods require a network forward pass for each sample (\ie $B=G$), the influence of $B$ on training cost is usually omitted in the literature. To align with previous approaches, we report an effective NFE per sample, calculated as $\text{NFE}_\theta =  \frac {B} {G} \cdot K$. In our experiments, we always use $G=12$.

Note that the training time per iteration in Table~\ref{tab:train_cost} shows a linear reduction as $\text{NFE}_\theta$ decreases. We emphasize that the policy update occupies the major part during the whole training, because the forward-backward calculation (with even slower gradient checkpoints) is much more costly than forward-only sampling. Our approach is demonstrated to be effective in reducing this part of cost from over 100 to 45 seconds per iteration. For a standard 300-iteration training, it takes about 4 hours on 32 NVIDIA H800 GPUs, which makes the GRPO training of flow models more practical.

\begin{figure}
    \centering
    \includegraphics[width=0.88\linewidth]{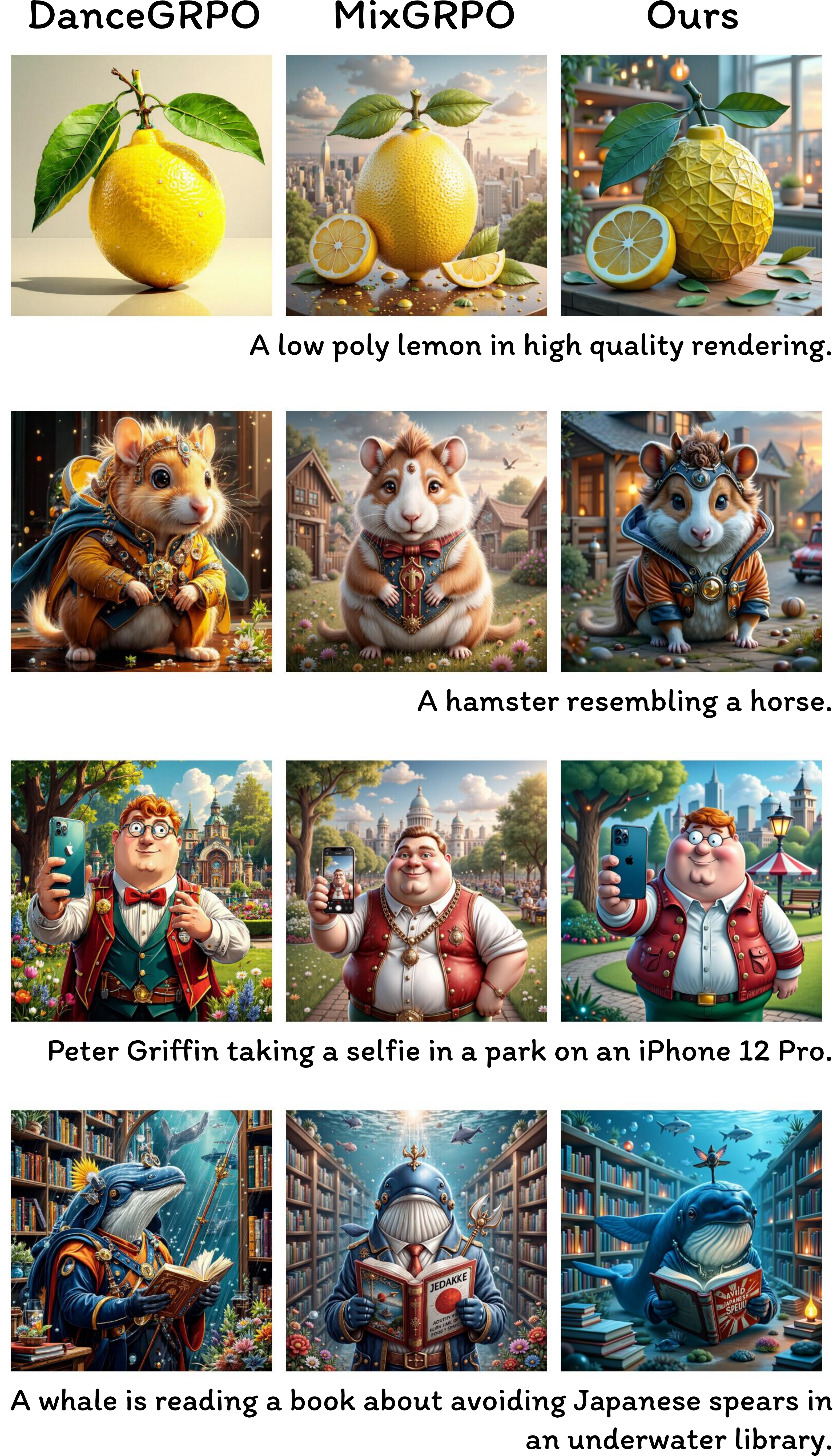}
    \caption{Visualization of different approaches.}
    \label{fig:vis_cmp}
\end{figure}

\vspace{-0.3cm}\paragraph{Qualitative Results.}
We present samples in Figure~\ref{fig:vis_cmp} and the supplementary materials. The images are generated with models trained with our approach (Ours with DPM++ solver and NFE$_\text{old}=8$ in Table~\ref{tab:major_cmp}) and the released baseline methods. Compared to baselines, our model better understands the user prompt and generates high-quality results without reward hacking artifacts like grid patterns and imbalanced color distribution. This visualization demonstrates the quantitative results in Table~\ref{tab:major_cmp}: our approach achieves better human preference alignment without hacking the in-domain reward models.

\begin{table}[t]
    \centering
\scalebox{0.85}{
    \begin{tabular}{lrrrrrr}
    \toprule
    \textbf{Method} & \textbf{NFE}$_{\text{old}}$ & $G$ & $B$ &  $K$  & \textbf{NFE}$_{\theta}$ & \textbf{s / Iter.}  \\
    \midrule
    DanceGRPO      & 25 & 12 & 12 &  14 & 14\textcolor{DEEPGREEN}{.00} & 237.86 \\
    MixGRPO        & 25 & 12 & 12 &  14 & 14\textcolor{DEEPGREEN}{.00} & 237.71\\
    \quad-Flash*   & 8 & 12 & 12 &  4 & 4\textcolor{DEEPGREEN}{.00} & 129.36 \\
    BranchGRPO     & 20 & 16 & adapt &  4 & 13.68 & 249.33 \\
    \midrule
    Ours 25-step  & 25 & 12 & 12 &  4 & 4\textcolor{DEEPGREEN}{.00} & 141.93 \\ 
    Ours 16-step  & 16 & 12 & 4 &  9 & 3\textcolor{DEEPGREEN}{.00} & 105.85 \\
    \rowcolor{SHALLOWGREEN} Ours 8-step  & 8 & 12 & 4 &  4 & 1.33 & 45.08\\ 
    
    \bottomrule
    \end{tabular}
}
    \caption{Training cost of different methods.}
    \label{tab:train_cost}
\end{table}

\newcommand{\vissubfig}[2]{{
\begin{subfigure}{0.25\linewidth}
        \centering
        \includegraphics[width=0.99\linewidth]{{#1}}
        \caption*{{#2}}
    \end{subfigure}
}}

\newcommand{\vissubfignocap}[1]{{
\begin{subfigure}{0.25\linewidth}
        \centering
        \includegraphics[width=0.99\linewidth]{{#1}}
    \end{subfigure}
}}


\begin{figure}
    \centering
    \includegraphics[width=0.95\linewidth]{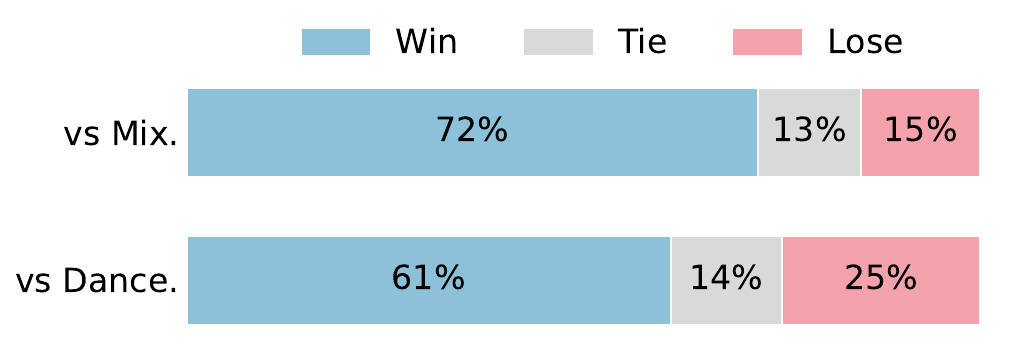}
    \caption{Human evaluation comparing our method with baselines.}
    \label{fig:humaneval}
\end{figure}

\vspace{-0.3cm}\paragraph{Human Evaluation.}
RLHF approaches are known to suffer from the reward hacking issue~\cite{xue2025dancegrpo}. Therefore, to further demonstrate our approach, we conduct a user study to compare the win rates against the baselines. We ask 46 users covering various professional backgrounds to choose the preferred image from each pair generated from the same HPDv2 prompt using different models. As shown in Figure~\ref{fig:humaneval}, our approach achieves higher human preference ratings (72\% and 61\%) than the baselines on the evaluated text-to-image generation tasks.

\subsection{Ablation Study}
We conduct a series of ablation studies to validate the effectiveness and parameter choices of the core components of Neighbor GRPO: the randomness strength $\sigma$ of ODE neighborhood sampling, the batch size $B$ of symmetric anchor sampling strategy, and the quasi-norm reweighting parameter $p$. We set $(\sigma, B, p) = (0.5, 4, 2)$ as the initial baseline for ablation and study around this default setup.

\newcommand{\visneighborssubfig}[1]{{
\begin{subfigure}{0.22\linewidth}
        \includegraphics[width=1\linewidth]{{#1}}
    \end{subfigure}
}}


\begin{figure}[t]
    \centering
    \includegraphics[width=0.9981\linewidth]{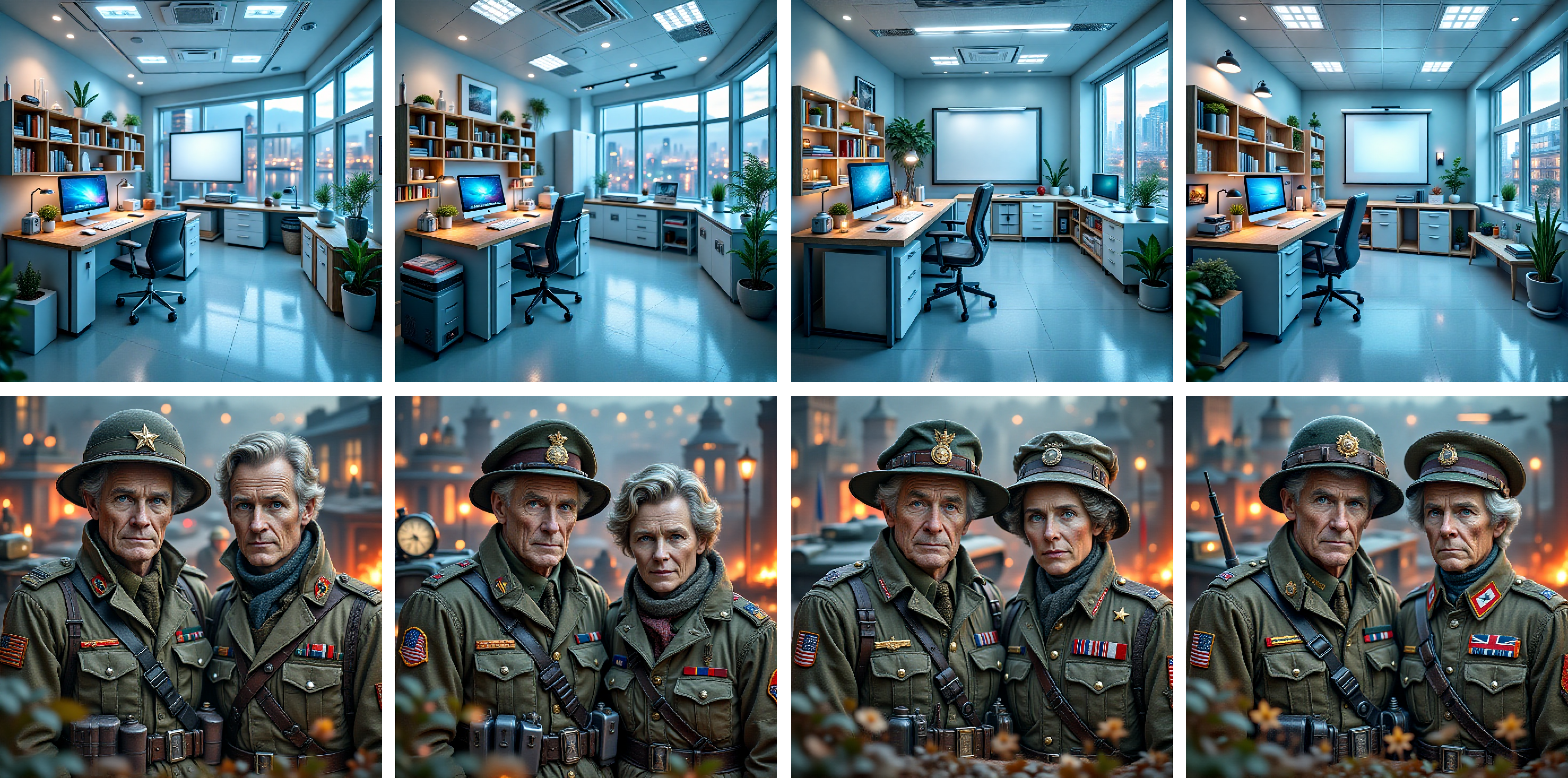}
    \caption{Rollout groups after 300-iteration training. $\sigma=0.3$.}
    \label{fig:visneighbors}
\end{figure}

\vspace{-0.5cm}\paragraph{Impact of ODE Neighborhood Sampling.}
To verify the necessity of our proposed neighborhood sampling strategy, we analyze the impact of the initial noise perturbation strength $\sigma$. As shown in Table~\ref{tab:randomness_cmp}, we experiment with different values for $\sigma$ in $\{0.1, 0.3, 0.5, 1.0\}$. We expect to find an optimal range; a value that is too small would create a neighborhood that is too tight for effective exploration, while a value that is too large would make the contrastive learning task too difficult and raise the risk of reward hacking, as the samples would no longer be considered neighbors. As the results show in Table~\ref{tab:randomness_cmp}, a larger $\sigma$ leads to better in-domain performance but may result in worse out-of-domain performance. We find that $\sigma=0.3$ is a suitable choice. Figure~\ref{fig:visneighbors} shows the sample diversity after sufficient training.

\vspace{-0.5cm}\paragraph{Symmetric Anchor Sampling.}
We investigate the influence of the number of anchors for each GRPO iteration. Table~\ref{tab:anchormerge} presents the results. With only $B=2$ anchors sampled, the HPSv2.1 score achieves over 0.37, which is a competitive performance. Note that in this situation, the policy-update process is 6 times faster. With a larger $B=4$ setup, all performance metrics improved. We find this to be a good balance between performance and training cost.

\vspace{-0.5cm}\paragraph{Group-Wise Quasi-Norm Reweighting.}
We validate our proposed Group-wise Quasi-Norm Reweighting by comparing it against the standard advantage normalization used in GRPO. See Table~\ref{tab:pnorm_cmp}.
We experiment with different values of $p$ in our $L_p$-norm, such as $\{0.5, 0.8, 1, 2\}$. The case where $p=2$ recovers the performance of standard group normalization. The results confirm that a value of $p=0.8$ is more effective against reward hacking, yielding the best out-of-domain scores. When adopting a smaller $p$ value, the model lacks exploration, as indicated by a decrease across all scores. The quantitative results demonstrate that the choice of a non-Euclidean norm is a critical factor.

\begin{table}[t]
    \centering
\scalebox{0.9}{
    \begin{tabular}{lrrrrrr}
    \toprule
    \textbf{$\sigma$}  &  \underline{\textbf{HPSv2.1}}  & \textbf{CLIP} & \textbf{Unified.} & \textbf{Aes.}  \\
    \midrule
    1.0 & 0.364 & 0.359 & 3.046 & 6.362 \\  
    0.5 & \textbf{0.379} & \textbf{0.368}  & 3.132 & 6.503 \\ 
    \rowcolor{SHALLOWGREEN} 0.3 & 0.375 & 0.363 & \textbf{3.141} & \textbf{6.639}   \\ 
    0.1 & 0.357 & 0.364 & 3.122 & 6.295 \\   
    \bottomrule
    \end{tabular}
}
    \caption{Performance with various perturbation strength.}
    \label{tab:randomness_cmp}
\end{table}
\begin{table}[t]
    \centering
\scalebox{0.9}{
    \begin{tabular}{lrrrrrr}
    \toprule
    \textbf{$B$}  &  \underline{\textbf{HPSv2.1}}  & \textbf{CLIP} & \textbf{Unified.} & \textbf{Aes.}  \\
    \midrule
    2 & 0.370 &  0.363 & 3.081 & 6.319\\
    \rowcolor{SHALLOWGREEN} 4 & \textbf{0.379} & 0.368  & 3.132 & \textbf{6.503} \\ 
    8 & 0.374 &  \textbf{0.373} & \textbf{3.187} & 6.403\\ 
    12 & 0.377 & 0.370 & 3.185&6.328\\ 
    \bottomrule
    \end{tabular}
}
    \caption{Performance of various number of sampled anchors.}
    \label{tab:anchormerge}
\end{table}
\begin{table}[t]
    \centering
\scalebox{0.9}{
    \begin{tabular}{lrrrrrr}
    \toprule
    \textbf{$p$}  &  \underline{\textbf{HPSv2.1}}  & \textbf{CLIP} & \textbf{Unified.} & \textbf{Aes.}  \\
    \midrule
    2 (GRPO)  & \textbf{0.379} & 0.368  & 3.132 & 6.503 \\ 
    1& 0.377 & 0.356 & 3.126 & 6.508    \\        
    \rowcolor{SHALLOWGREEN} 0.8& 0.372 & \textbf{0.371}  & \textbf{3.166} &  \textbf{6.626} \\   
    0.5& 0.372 & 0.354 & 3.104 &  6.548 \\  
    \bottomrule
    \end{tabular}
}
    \caption{Performance with various $p$ normalization.}
    \label{tab:pnorm_cmp}
\end{table}

\section{Conclusion}

In this paper, we introduced Neighbor GRPO, a novel RLHF paradigm for aligning flow matching models that completely bypasses the need for SDE conversion. Our core insight is to reinterpret the optimization dynamics of existing SDE-based GRPO methods as a form of distance-based contrastive learning. Based on this, we integrated a surrogate leaping policy into the GRPO framework, which provides both the necessary stochastic abstraction and contrastive learning dynamics. Furthermore, we assess practical implementations.

Our extensive experiments demonstrated that Neighbor GRPO significantly outperforms SDE-based counterparts. It fully preserves the advantages of deterministic ODE sampling, including compatibility with high-order solvers. The method achieves faster convergence and superior generation quality, highlighting the effectiveness of its more direct credit assignment mechanism.


\section*{Acknowledgement}

This study was supported in part by National Key R\&D Program of China Project 2022ZD0161100, in part by the Centre for Perceptual and Interactive Intelligence, a CUHK-led InnoCentre under the InnoHK initiative of the Innovation and Technology Commission of the Hong Kong Special Administrative Region Government, in part by NSFC-RGC Project N\_CUHK498/24, and in part by Guangdong Basic and Applied Basic Research Foundation (No.2023B1515130008, XW).
    
{
    \small
    \bibliographystyle{ieeenat_fullname}
    \bibliography{main}
}


\end{document}


\maketitle

\section{Algorithm}

We provide the pseudo-code for Neighbor GRPO in Algorithm~\ref{alg:neighbor_grpo_rollout}. We suggest adopting a more efficient solver, such as DPM-Solver++, with fewer sampling timesteps $T$ for the $\mathrm{RolloutSolver}$ to accelerate training. Compared with SDE-based GRPO methods, which call $\mathrm{TrainSolver}$ $G\times K$ times in each GRPO iteration, our approach requires $B\times K$ calls to $\mathrm{TrainSolver}$. Note that we usually have $B<G$, which allows us to further reduce the training cost with Neighbor GRPO.

\begin{algorithm}[t]
  \caption{Neighbor GRPO}
  \label{alg:neighbor_grpo_rollout}
  \begin{algorithmic}
  \Require {prompts $C$, parameters $\theta$; group size $G$; anchors per batch $B$; number of rollout steps $T$; number of train steps $K$; time schedule $\mathcal{T} = \{t_1, t_2, \dots, t_T\}$}
  \Ensure {$t_T = 1$, $t_1 = 0$}
  \For{iteration $m=1$ to $M$}
    \State $\pi_{\theta_\text{old}} \gets \pi_{\theta}$
    \State Sample prompt $c\in C$
    \State Draw shared initial noise $\epsilon^\ast \sim \mathcal{N}(0,I)$
        \For{$i = 1$ to $G$}
              \State Draw $\delta^{(i)} \sim \mathcal{N}(0,I)$
              \State $x^{(i)}_{1} \leftarrow \sqrt{1-\sigma^2}\epsilon^\ast + \sigma \delta^{(i)}$ \Comment{Perturb noise.}
              \For{$k = T$ to $1$}{
                \State $x^{(i)}_{t_{k-1}} \leftarrow \mathrm{RolloutSolver}.\mathrm{step}\big(x^{(i)}_{t_k}, c; \theta_{\text{old}}\big)$
                \State\Comment{Stop gradient.}
              \EndFor
        \EndFor
            
        \State Compute advantages $\{A_i\}$
        \State Sample $B$ anchor indices $S \subset \{1,\dots,G\}$
        \State Sample $K$ steps $\mathcal{K} \subset \{1, 2, \dots, T-1\}$
        \For{$j \in S$}
          \For{$k \in \mathcal{K}$}
            \State $x^{(\theta)}_{t_{k+1}} \leftarrow x^{(j)}_{t_{k+1}}$
            \State $x^{(\theta)}_{t_k} \leftarrow \mathrm{TrainSolver}.\mathrm{step}\big(x^{(\theta)}_{t_{k+1}}, c; \theta\big)$
            \For {$i=1$ to $G$ }
                \State $\pi_\theta^{(i)} \leftarrow \mathrm{softmax}_i(-\|x^{(i)}_{t_k} - x^{(\theta)}_{t_k}\|_2^2)$ 
                \State $\pi_{\theta_\text{old}}^{(i)} \leftarrow \mathrm{softmax}_i\!\big(-\|x^{(i)}_{t_k}-x^{(j)}_{t_k}\|_2^2\big)$ 
                \State $\rho^{(i\mid j)}\gets \pi_{\theta}^{(i)}/\pi_{\theta_\text{old}}^{(i)}$
            \EndFor
            \State Compute $J = J(A, \rho; \theta)$ 
            \State Accumulate gradients $\nabla_\theta J$
          \EndFor
          \State Update $\theta$ using accumulated gradients
        \EndFor
    \EndFor
  }
  \end{algorithmic}
\end{algorithm}

\section{Derivation of the SDE Formulation}
For a Rectified Flow model, the forward process from data $x_0$ to noise $\epsilon$ is $x_t = (1-t)x_0 + t\epsilon$. The reverse process follows the ODE $\frac{dx_t}{dt} = v_\theta(x_t, t)$, where $v_\theta(x_t,t) \approx \epsilon - x_0$.

To introduce stochasticity, we construct an equivalent SDE that shares the same marginal probability density $p_t(x)$ for all $t \in [0,1]$. A common choice is:
\begin{equation}
  dx_t = \left[ v_\theta(x_t,t) + \frac{\eta_t^2}{2t} \hat{\epsilon}_\theta(x_t,t) \right] dt + \eta_t dw_t,
\end{equation}
where $dw_t$ is a standard Wiener process, $\eta_t$ controls the noise intensity, and $\hat{\epsilon}_\theta(x_t,t) = x_t + (1-t)v_\theta(x_t,t)$ is the estimated initial noise.

For numerical simulation, we discretize the reverse SDE. With a backward step $\Delta t > 0$, the update rule is:
\begin{equation}
  x_{t-\Delta t} = x_t - \left[ v_\theta(x_t,t) + \frac{\eta_t^2}{2t}\hat{\epsilon}_\theta(x_t,t) \right] \Delta t + \eta_t \sqrt{\Delta t} \epsilon,
\end{equation}
where $\epsilon \sim \mathcal{N}(0,I)$.
Let's define the single-step ODE update as $x_{t-\Delta t}^{(\mathrm{ODE})} = x_t - v_\theta(x_t,t)\Delta t$. Let $\sigma_t^2 = \eta_t^2\Delta t$. The SDE update can be rewritten as:
\begin{align}
  x_{t-\Delta t}^{(\mathrm{SDE})} &= x_t - v_\theta(x_t,t)\Delta t - \frac{\sigma_t^2}{2t\Delta t}\hat{\epsilon}_\theta(x_t,t)\Delta t + \sigma_t \epsilon \\
  &= x_{t-\Delta t}^{(\mathrm{ODE})} - \frac{\sigma_t^2}{2t}\hat{\epsilon}_\theta(x_t,t) + \sigma_t \epsilon.
\end{align}

This discretized update defines a Gaussian policy $\pi_\theta(x_{t-\Delta t} | x_t, c) = \mathcal{N}(\mu_\theta, \sigma_t^2 I)$, where the mean is $\mu_\theta(x_t,t) = x_{t-\Delta t}^{(\mathrm{ODE})} - \frac{\sigma_t^2}{2t}\hat{\epsilon}_\theta(x_t,t)$. The negative log-likelihood of this policy is:
\begin{equation}
  -\log \pi_\theta(x_{t-\Delta t}\mid x_t,c) = \frac{1}{2\sigma_t^2}\| x_{t-\Delta t} - \mu_\theta(x_t,t)\|_2^2 + \text{const}.
\end{equation}
This provides the basis for the distance-based contrastive learning interpretation of SDE-based GRPO.

\section{Derivation of the Drift Residual}

The objective of GRPO is to maximize the expected log-likelihood of samples from the old policy $\pi_{\theta_{\text{old}}}$ under the new policy $\pi_\theta$. This is equivalent to minimizing the negative log-likelihood:
\begin{equation}
  \mathcal{L}(\theta) = \mathbb{E}_{x_{t-\Delta t} \sim \pi_{\theta_{\text{old}}}} [-\log \pi_\theta(x_{t-\Delta t})].
\end{equation}

From the SDE formulation in the previous section, the policy $\pi_\theta$ is an isotropic Gaussian distribution $\pi_\theta(x_{t-\Delta t} | x_t) = \mathcal{N}(\mu_\theta, \sigma_t^2 I)$. The mean $\mu_\theta$ and a sample from the old policy $x_{t-\Delta t}^{(\mathrm{SDE, old})}$ are given by:
\begin{align}
  \mu_\theta &= x_{t-\Delta t}^{(\mathrm{ODE}, \theta)} - \frac{\sigma_t^2}{2t}\hat{\epsilon}_\theta(x_t,t), \\
  x_{t-\Delta t}^{(\mathrm{SDE, old})} &= x_{t-\Delta t}^{(\mathrm{ODE, old})} - \frac{\sigma_t^2}{2t}\hat{\epsilon}_{\theta_\text{old}}(x_t,t) + \sigma_t \epsilon,
\end{align}
where $\epsilon \sim \mathcal{N}(0,I)$, and the superscripts denote the policy parameters used for the ODE step and noise prediction.

The negative log-likelihood term, omitting constants, is a mean squared error (MSE):
\begin{equation}
  -\log \pi_\theta(x_{t-\Delta t}^{(\mathrm{SDE, old})}) = \frac{1}{2\sigma_t^2} \left\| x_{t-\Delta t}^{(\mathrm{SDE, old})} - \mu_\theta \right\|^2_2.
\end{equation}

Substituting the expressions for the mean and the sample, we get the term inside the norm:
\begin{align}
  & x_{t-\Delta t}^{(\mathrm{SDE, old})} - \mu_\theta \nonumber \\
  &= \left( x_{t-\Delta t}^{(\mathrm{ODE, old})} - \frac{\sigma_t^2}{2t}\hat{\epsilon}_{\theta_\text{old}} + \sigma_t \epsilon \right) - \left( x_{t-\Delta t}^{(\mathrm{ODE}, \theta)} - \frac{\sigma_t^2}{2t}\hat{\epsilon}_\theta \right) \\
  &= \left( x_{t-\Delta t}^{(\mathrm{ODE, old})} + \sigma_t \epsilon \right) - x_{t-\Delta t}^{(\mathrm{ODE}, \theta)} - \left( \frac{\sigma_t^2}{2t}\hat{\epsilon}_{\theta_\text{old}} - \frac{\sigma_t^2}{2t}\hat{\epsilon}_\theta \right).
\end{align}

The main paper presents this MSE term in a simplified form:
\begin{equation}
  \frac{1}{2\sigma_t^2} \left\| \tilde{x}_{t-\Delta t}^{(\mathrm{ODE})} - x_{t-\Delta t}^{(\mathrm{ODE})} + o_t(x_t) \right\|^2_2,
\end{equation}
where $\tilde{x}_{t-\Delta t}^{(\mathrm{ODE})} = x_{t-\Delta t}^{(\mathrm{ODE, old})} + \sigma_t\epsilon$ is the perturbed ODE sample from the old policy, and $x_{t-\Delta t}^{(\mathrm{ODE})}$ represents the ODE step from the new policy.

By comparing the two forms, we can identify the drift residual term $o_t(x_t)$:
\begin{align}
  o_t(x_t) 
  &= \frac{\sigma_t^2}{2t}\hat{\epsilon}_\theta(x_t,t) - \frac{\sigma_t^2}{2t}\hat{\epsilon}_{\theta_\text{old}}(x_t,t).
\end{align}
This term represents the drift correction between the old and new policies. As the policy update is constrained within a trust region, $\theta \approx \theta_{\text{old}}$, and this residual term approaches zero.

\section{Marginal Distribution Preservation}
\label{sec:marginal}

Consider the standard flow matching framework with marginal distribution $p_\theta(x_t)$ at timestep $t \in [0, 1]$, induced by the deterministic ODE flow $\Phi_{1 \to t}(\cdot; \theta)$ starting from initial noise $x_1$:
\begin{equation}
    p_\theta(x_t) = \mathbb{E}_{x_1\sim\mathcal{N}(0;I)}\left[ \delta(x_t - \Phi_{1 \to t}(x_1; \theta))\right],
\end{equation}
where $\delta(\cdot)$ is the Dirac delta function. We define a leaping probability $\pi(x_t | x_t') = \mathcal{N}(x_t | x_t', \sigma_\pi I)$, denoting the probability that any reachable latent sample $x_t' = \Phi_{1 \to t}(x_1'; \theta)$ (derived from an initial noise $x_1'$) leaps to $x_t$. Following this distribution, the marginal distribution of $x_t$ leaped from any $x_t'$ is
\begin{equation}
    \pi(x_t) = \int p_\theta(x_t') \pi(x_t | x_t') dx_t'.
\end{equation}
Given a sufficiently small $\sigma_\pi$ (or as $\sigma_\pi \to 0$), the support of $\pi(x_t | x_t')$ collapses to a single point where $x_t' = x_t$. Consequently, $\pi(x_t) \to p_\theta(x_t)$.

Now consider a Monte-Carlo approximation of $\pi(x_t)$:
\begin{equation}
    \hat{\pi}(x_t) = \frac{1}{G} \sum_{i=1}^{G} \pi(x_t|x_t^{(i)}),  \quad {x_t^{(i)}\sim p_\theta(x_t)}.
\end{equation}
Let $\hat{\pi}(x_t^{(i)}) = \frac{1}{G}$. We have:
\begin{align}
    \hat{\pi}(x_t^{(i)} | x_t) &= \frac{\hat{\pi}(x_t^{(i)}) \hat{\pi}(x_t\mid x_t^{(i)}) }{\hat{\pi}(x_t)}
    \\ &= \frac{ \pi(x_t|x_t^{(i)})} {\sum_{j=1}^{G} \pi(x_t|x_t^{(j)})} \\
    &=\frac{\exp(-\|x_t - x_t^{(i)}\|^2_2 / 2\sigma_\pi^2)}{\sum_j\exp(-\|x_t - x_t^{(j)}\|^2_2 / 2\sigma_\pi^2)}.
\end{align}
This formulation results in a softmax distribution with a temperature of $2\sigma^2_\pi$. In practice, we omit this temperature parameter because the Euclidean distance $\|x_t - x_t^{(i)}\|^2_2$ is empirically large enough when $i \neq j$. Replacing $x_t$ with $x_t^{(\theta)}$ yields the surrogate leaping policy presented in the main paper. Maximizing this posterior probability $\hat{\pi}(x_t^{(i)} | x_t^{(\theta)})$ encourages the generated sample to align closely with the support of the empirical marginal distribution defined by $\{x_t^{(i)}\}_{i=1}^G$. This serves as a tractable surrogate objective for marginal distribution preservation.

\section{More Qualitative Results}

\begin{figure*}
    \centering
    \includegraphics[width=0.98\linewidth]{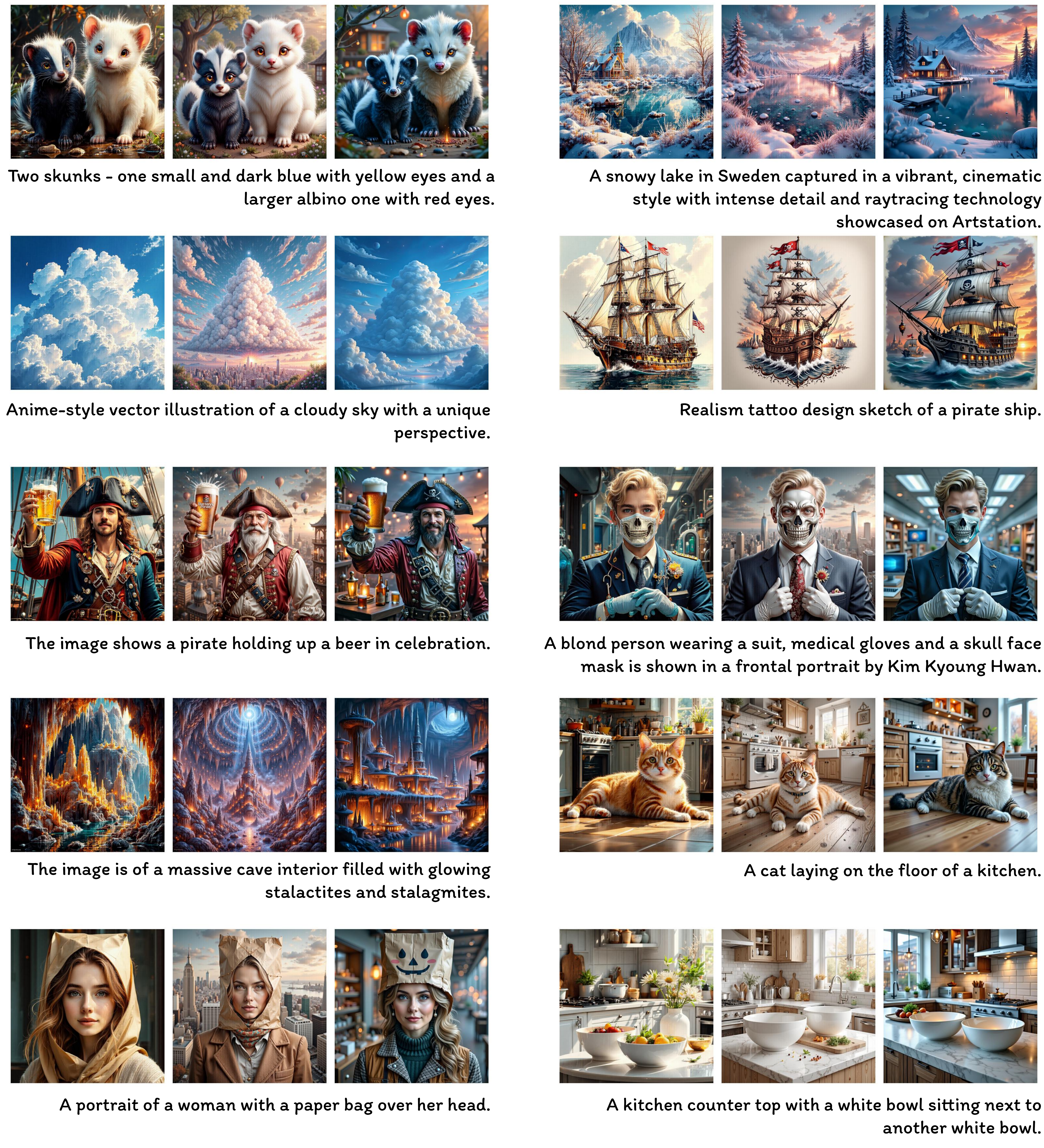}
    \caption{Visualization of different approaches (DanceGRPO - MixGRPO - NeighborGRPO).}
    \label{fig:supp_vis}
\end{figure*}

We present more visualization results comparing different methods in Figure~\ref{fig:supp_vis}.

